\documentclass[letterpaper, 10 pt, journal, twoside]{IEEEtran}

\IEEEoverridecommandlockouts                              

\pdfoutput=1

\usepackage{graphicx}
\graphicspath{ {fig/} }
\usepackage{subcaption}
\usepackage{amssymb}
\usepackage{amsmath,amsfonts}
\usepackage{mathtools}
\usepackage{caption}
\usepackage{hyperref}
\usepackage{comment}
\usepackage{multirow}
\usepackage{graphicx}
\usepackage{adjustbox}
\usepackage{mathtools}
\usepackage{wrapfig}
\DeclarePairedDelimiterX{\infdivx}[2]{(}{)}{%
  #1\;\delimsize\|\;#2%
}

\usepackage[dvipsnames]{xcolor}
\newcommand{\edits}[1]{\textcolor{black}{#1}}
\usepackage{algorithm}
\usepackage{algorithmic}
\usepackage{array}
\usepackage{makecell}

\usepackage[top=54pt, left=54pt, right=54pt, bottom=54pt]{geometry}

\begin{document}

\title{
Diffusion Co-Policy for Synergistic Human-Robot Collaborative Tasks
}

\author{Eley Ng$^{1}$, Ziang Liu$^{2}$, and Monroe Kennedy III$^{1,2}$
\thanks{Manuscript received: May 20, 2023; Revised:
August 16, 2023; Accepted: October 16, 2023.}
\thanks{This paper was recommended for publication by
Editor Angelika Peer upon evaluation of the Associate Editor and Reviewers’
comments. The first author is supported by the NSF Graduate Research Fellowship. This work was supported by Stanford Institute for Human-Centered Artificial Intelligence (HAI) and conducted under IRB-65022. Link to project site: \href{https://sites.google.com/view/diffusion-co-policy-hrc}{https://sites.google.com/view/diffusion-co-policy-hrc}.}
\thanks{$^{1}$Department of Mechanical Engineering, Stanford University, USA. 
{\tt\small \{eleyng, monroek\}@stanford.edu.} }
\thanks{$^{2}$Department of Computer Science, Stanford University, USA. 
{\tt\small \{ziangliu\}@stanford.edu.} }
\thanks{Digital Object Identifier (DOI): see top of this page.}
}

\markboth{IEEE ROBOTICS AND AUTOMATION LETTERS. PREPRINT VERSION. ACCEPTED
OCTOBER, 2023}
{Ng \MakeLowercase{\textit{et al.}}: Diffusion Co-Policy for Synergistic Human-Robot Collaborative Tasks} 


\maketitle

\begin{abstract}
Modeling multimodal human behavior has been a key barrier to increasing the level of interaction between human and robot, particularly for collaborative tasks. \edits{Our \textbf{key insight} is that an effective, learned robot policy used for human-robot collaborative tasks must be able to express a high degree of multimodality, predict actions in a temporally consistent manner, and recognize a wide range of frequencies of human actions in order to seamlessly integrate with a human in the control loop. We present Diffusion Co-policy, a method for planning sequences of actions that synergize well with humans during test time. The co-policy predicts joint human-robot action sequences via a Transformer-based diffusion model, which is trained on a dataset of collaborative human-human demonstrations, and directly executes the robot actions in a receding horizon control framework.} We demonstrate in both simulation and real environments that the method outperforms other state-of-art learning methods on the task of human-robot table-carrying with a human in the loop. Moreover, we qualitatively highlight compelling robot behaviors that demonstrate evidence of true human-robot collaboration, including mutual adaptation, shared task understanding, leadership switching, and low levels of wasteful interaction forces arising from dissent. 

\end{abstract}


\section{Introduction}
\IEEEPARstart{M}{ultimodal} behavior poses a key barrier to achieving effective human-robot coordination in collaborative tasks. In collaborative tasks, decentralized agents execute joint actions, which are defined in cognitive neuroscience as ``any form of social interaction whereby two or more individuals coordinate their actions in space and time to bring about a change in the environment" \cite{sebanz2006joint}. For such scenarios, the ability to anticipate and predict partners' actions is crucial, as it can significantly enhance the capacity to plan synergistic actions that contribute to the team's success with an understanding of the task (i.e. how the team's actions affects the dynamics). Collaborative table carrying, an exemplar of such tasks, demands on-the-fly mutual adaptation and precise timing of movements. This paper introduces a new approach to coordinating and executing behaviors seamlessly, demonstrated through human-robot collaborative carrying.

\par Given the recent successes of denoising diffusion probabilistic models (DDPM) in learning single agent behaviors \cite{janner2022planning}, \edits{\cite{urain2023se}}, \cite{chi2023diffusion}, we propose leveraging DDPMs in human-robot collaboration, and demonstrate that the gains in model expressivity enabled by DDPM are highly suited for tasks that involve humans. Using human-human collaborative demonstrations, we train a robot co-policy that conditions on past observations and human actions to generate sequences of future joint human and robot actions, and directly execute the robot actions using receding horizon control. We show the effectiveness of the diffusion-based co-policy in both simulation and real robot experiments by highlighting compelling collaborative behaviors exhibited by the robot and human. 

\begin{figure}[t]
  \centering
  \includegraphics[width=\linewidth]{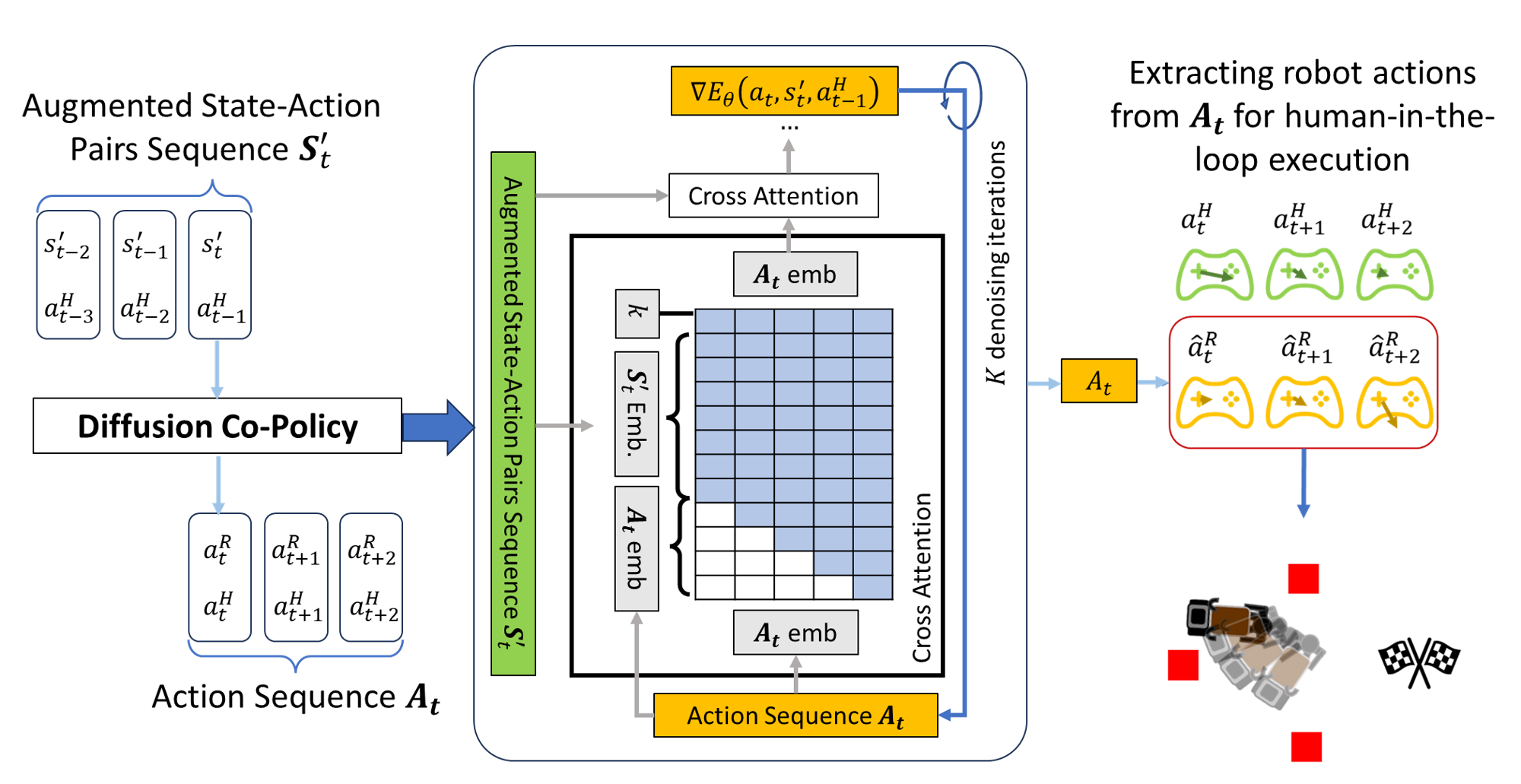}
  \vspace{-5mm}
  \caption{ \edits{Framework for human-robot table-carrying with diffusion co-policy. At time step $t$, the co-policy takes as input the latest $T_o$ steps of augmented state data and past human actions, denoted as the augmented state-action pair sequence $S'_t$, and outputs $T_a$ steps of joint actions $A_t$. To generate the output, the Transformer-based decoder architecture denoises a sampled action sequence after $K$ iterations through multiple cross-attention layers. The robot extracts the robot action sequence from $A_t$ and executes it simultaneously with control inputs from a human.}}
  \label{fig:maps}
  \vspace{-3mm}
  
\end{figure}
        
\par \textbf{Contributions}: Our primary contribution is the application of diffusion models for learning robot collaborative policies that can synergize with a real, human partner. Our approach utilizes a Transformer-based diffusion model to predict future joint action sequences conditioned on past observations and human actions, facilitating smooth, coordinated actions that can be executed without further processing or use of hand-engineered collaborative reward functions. We demonstrate on the table-carrying task that this method outperforms state-of-the-art imitation learning approaches in both simulated and real-world experiments with humans, achieving higher task success rates and lower, wasteful interaction forces. By leveraging the generative capabilities of diffusion models, we have presented a significant step towards enabling effective human-robot collaboration on continuous state, joint-action tasks that require rapid mutual co-adaptation, behavior multimodality, coordination, and shared task understanding via learned, implicit constraints from human interaction.

\section{Related Work}
\par \textbf{Imitation learning}: A multitude of work has been done in improving the quality of policies trained with offline datasets, particularly to account for multimodality. In particular, behavioral cloning has made significant progress due to advances \edits{in policy representation, by moving from \textit{explicit} representations (e.g. LSTM-GMM \cite{mandlekar2021matters}, Transformers \cite{shafiullah2022behavior}, normalizing flows \cite{urain2020imitationflow}) to \textit{implicit} representations (energy-based models \cite{pmlr-v164-florence22a}, score-based models \cite{song2019score}, diffusion models \cite{chi2023diffusion}), which allow for the expression of multimodal outcomes. Chi, et. al. showed that the diffusion policy representation can surpass other policy representations in expressing multimodality over a variety of tasks \cite{chi2023diffusion}}. While these methods demonstrate the generative power of diffusion models, their effectiveness in scenarios that involve human behavior recognition remains to be seen. 


\par \textbf{Human intent modelling for HRI}: Prior work\edits{s} in human-robot interaction (HRI) have generally fallen along a spectrum defined by the degree to which the human or the world is modelled. Theory of Mind (ToM) methods, which ascribe mental states to the human with whom the robot interacts, generally involve learning human reward functions \cite{baker2009action, sadigh2016planning}, learning user type \cite{nikolaidis2015efficient}, \edits{human motion prediction \cite{chalvatzaki2019learn},} or latent strategies \cite{xie2021learning}. These approaches are predicated on the hypothesis that human behavior follows a goal-directed policy, but are not necessarily structured to allow for multimodal behaviors given an attribute. \edits{Contrary to ToM methods,} black box methods leverage data to \edits{directly train} robot policies, though many approaches are a mix of both. Recent advances in HRI suggest that a promising approach for modelling multimodal behavior is to leverage the expressiveness of generative models for planning (e.g. variational autoencoders \cite{schmerling2018multimodal, butepage2020imitating, prasad2022mild}, variational recurrent neural networks \cite{ng2022takes}), or learning a joint action policy (e.g. Co-GAIL \cite{wang2022co}). 

\par In this work, we leverage diffusion models to learn a co-policy, wherein we predict future action sequences of both agents given sequences of past partner actions and observations. Nikolaidis, et. al. \cite{nikolaidis2017human} similarly investigates the human-robot table carrying task, using a discrete state-action formulation to model human adaptation during interaction. While our continuous state-action formulation does not explicitly model adaptation, it enables the robot to execute dynamic behaviors alongside a human in real-time. 


\section{Diffusion Co-Policy Formulation}
This section describes the formulation of the collaborative robot policy as a \edits{DDPM}. First, we describe the collaborative task and motivate the use of diffusion models, leading to the formulation of a co-policy which incorporates human action and scene conditioning for significantly improved human-robot coordination on continuous state-action tasks. 

\subsection{Problem Setting}
\par Consider a human-robot system wherein the dynamics of the world state $s \in \mathcal{S} \subset \mathbb{R}^n$. \edits{Let $a^i \in \mathcal{A} \subset \mathbb{R}^m, i\in [H,R]$ define the action space of the human $H$ and robot $R$. At any point in time $t$, the human can take action $a_t^H$ jointly with the robot action $a_t^R$, and the joint action is denoted by the concatenated vector $\textbf{a}_t = (a_t^H, a_t^R)$. The state progresses with the following dynamics:}
$$
    s_{t+1} = f(s_t, \textbf{a}_{t}) \eqno{(1)} \\
$$
Provided with a control sequence of joint actions, a rollout starting from initial state $s_0$ results in trajectory $\tau = (s_0, \textbf{a}_0, ... s_T, \textbf{a}_T)$.

\par Typically, the goal is to find a sequence of actions $\textbf{a}^{*}_{0:T}$ that maximizes the sum of rewards $\sum^T_{t=0}r(s_t, \textbf{a}_t)$ via trajectory optimization or reinforcement learning methods. 
\par An instance of such a system is the collaborative carrying task \cite{ng2022takes}, a human and a robot both carry a table at opposite ends, moving it from a start pose to a goal location while avoiding obstacles. \edits{In this task, we assume full observability of the state. The state is a 7-dim. vector of the 2D table pose $(p_x, p_y, \theta)$ and its velocity in the world frame: $s = [p_x, \ p_y, \ cos\theta, \ sin\theta, \ \dot{p_x}, \ \dot{p_y}, \ \dot{\theta}].$ Furthermore, the state can be \textit{augmented} by concatenating the following: the initial pose of the table, the 2D goal position of the table, and a 6-dim observation of the map, which consists of a vector concatenation of the world-frame positions of a maximum of three obstacles in the map. We refer to this 18-dim augmented state as $s'$. Perception was not the focus of this work, so we used a low-dimensional vector representation; however, future work can amend this representation to an arbitrary number of local obstacles, or leverage visual information. Each agent's action is a 2D force applied at opposite ends of the table; thus, the joint action is 4-dim. } 

\subsection{Approach}
While we could formulate the diffusion model for planning with RL using classifier-guided sampling \cite{janner2022planning, dhariwal2021diffusion}, doing so would require hand-designing a collaborative reward function. However, manually designing reward functions \edits{is} tricky and prone to over-specification, particularly for multi-agent collaborative scenarios, where multimodal behaviors arise from preference and various factors like diversity and inconsistency. Despite promising directions in increasing the nuance of learned reward functions \cite{myers2022learning}, querying methods would not be viable in interactive, long-horizon tasks like collaborative carrying, as specifying the query itself would be non-trivial.

\edits{Our \textbf{key insight} is that an effective, learned robot policy used for human-robot collaborative tasks must be able to express a high degree of multimodality, predict actions in a temporally consistent manner, and recognize a wide range of frequencies of human actions in order to seamlessly integrate with a human in the control loop.} Given recent empirical breakthroughs in the generative quality of diffusion models \cite{ho2020denoising, song2020denoising}, we propose leveraging diffusion models to enable coordination on long-horizon, continuous state-action, human-robot collaborative tasks in novel settings during test time by relying on demonstrations to capture interaction dynamics.

\subsection{Denoising Diffusion Probabilistic Models}

DDPMs \cite{welling2011bayesian, sohl2015deep, ho2020denoising} are generative models that approach sample generation with an iterative denoising process modeled by Langevin dynamics. Data generation via diffusion works by denoising (reversing) a forward diffusion process that iteratively adds noise to data until it resembles as standard Gaussian. Specifically, samples from a standard Gaussian prior, $p(x_K) = \mathcal{N}(0, I)$ pass through $K$ iterations of noise reduction based on a fixed iteration-dependent variance schedule (parameterized by $\sigma_k$, $\alpha_k$ and $\gamma_k$), producing $K$ intermediate latent variables, $x_{k-1}, ..., x_0$, where $x_0$ is the noiseless output. The Gaussian noise predicted is parameterized by a network, $\epsilon_{\theta}(x_k, k)$. Thus, to sample $x_{k-1} \sim p(x_{k-1} | x_{k})$, we compute:

$$
    x_{k-1} = \alpha_k\Big(x_k - \gamma_k\epsilon_\theta(x_k, k)\Big) + \sigma_kz  \eqno{(2)} \label{eqn:ddpm_sampling}
$$
where $z \sim \mathcal{N}(0, I)$.

\par \textit{Clarification of notation:} In this work, we use two different time steps in subscript: $k$ to denote the diffusion timestep, and $t$ to denote the prediction timestep, i.e. $s_{t,k}$ is the $t^{th}$ state in the $k^{th}$ diffusion step. Subscripts of noiseless quantities are omitted, e.g. $s_t$. Subscripts of constants parameterized only by $k$ do not have a time-indexed subscript, e.g. $\epsilon_k$. 

\subsection{Diffusion Co-Policy for Coordinating with Humans}

\par We consider the task of modeling a robot co-policy as learning a probabilistic model for the robot that infers future \textit{sequences} of \textit{joint human-robot} actions, conditioned on past states, map information, and past human actions. \edits{ More specifically, we seek to model the conditional distribution $p(\textbf{a}_t | s'_t, a_{t-1}^H)$. Thus,} we modify the DDPM in two ways: 1. Conditioning on past human actions, which allows the robot to derive an understanding of human strategy from past human action trajectories to aid future team predictions; and 2. High-level goal conditioning, wherein the robot can condition its predictions on where the carried table should land. \edits{Eq. 2 can be modified to model the conditional distribution:}
$$
    \edits{\textbf{a}_{t, k-1}=  \alpha_{k}\Big(\textbf{a}_{t, k} - \gamma_k\epsilon_\theta(\textbf{a}_{t,k}, s'_t, a^H_{t-1}, k)\Big) + \sigma_k z}  \eqno{(3)}
$$ 

\edits{Eq. 3 can also be interpreted as a noisy gradient descent step, with the gradient of the energy-based model (EBM), $\nabla E_{\theta}(a_{t}, s'_t, a^H_{t-1})$, and learning rate $\gamma$:}
$$
    \edits{\textbf{a}_{t} \leftarrow \textbf{a}_{t} - \gamma \nabla E_{\theta}(\textbf{a}_{t}, s'_t, a^H_{t-1})  } \eqno{(4)}
$$
\edits{In other words, $\epsilon_\theta(\textbf{a}_{t,k}, s'_t, a^H_{t-1}, k)$ predicts $\nabla E_{\theta}(\textbf{a}_{t}, s'_t, a^H_{t-1})$, which approximates the action-score gradient, i.e. $\nabla \log p_\theta(\textbf{a}_t | s'_t, a_{t-1}^H) $. Song, et. al. \cite{song2019score} provides further background on score-based models and this relationship, but we summarize the implications. By learning the parameters of the action-score gradient (thus, the distribution $p_\theta(\textbf{a}_t | s'_t, a_{t-1}^H)$), we can circumvent approximating the intractable normalization constant, a problem which pervades likelihood-based models by affecting training stability and limiting model expressivity \cite{song2019score}. By performing Stochastic Langevin Dynamics sampling on this gradient field, the network can express arbitrary, multimodal distributions, which is beneficial for learning behaviors in human-robot interactive tasks.}

\subsection{Network Architecture}

We adopt the Transformer-Based Diffusion architecture from Chi, et. al. \cite{chi2023diffusion}, which uses the minGPT \cite{brown2020language} transformer decoder model for action prediction, and modify the inputs as follows. \edits{The model takes as input a sequence of $T_o$ steps of augmented state-action pairs; more specifically, these pairs consist of augmented states $s'$ and past human actions $a^H$. The model outputs a sequence of $T_a$ joint action steps denoised by the diffusion model.} Noise-injected joint actions, $\textbf{a}_{t,k}$, are tokenized and passed to the transformer decoder, which uses a sinusoidal embedding to encode the $k^{th}$ diffusion step inputs as well as $k$, which is prepended as first token. Positional embedding is applied to conditional inputs, $s_t'$ and $a_{t-1}^H$, which are converted to a sequence before passed to the transformer decoder. The decoder then predicts the noise corresponding to each input in the time dimension for the $k^{th}$ iteration.  A causal attention mask constrains the attention of each action to itself and prior actions. The predicted joint action sequence is constructed only after the predicted noise is propagated through the noise scheduler following the reverse diffusion process.

\par As expected \cite{chi2023diffusion}, the 1D temporal CNN diffusion model does not work as well as the Transformer-based model for this task due to oversmoothing the action space. For the task we focus on, the actions are inertial forces. Depending on the user and playing style, people tend to apply short impulses due to damping and oscillations when pivoting around obstacles or correcting speed, making the Transformer-based architecture useful for the table-carrying task.

\subsection{Training}

To train the diffusion model, \edits{ unnoised joint action data, $\textbf{a}_{t,0}$, and a value of $k$ are randomly sampled, the latter of which is then used to sample noise $\epsilon_k$ with variance $\sigma_k$. $\epsilon_\theta(\textbf{a}_{t,k}, s'_t, a^H_{t-1}, k)$ then predicts the noise from the data sample (with the noise added). The loss for the noise model is:}
$$
    \mathcal{L} = \|\epsilon_k - \epsilon_{\theta}(s'_t, a^H_{t-1}, \textbf{a}_{t,0} + \epsilon_k, k)) \|_2^2 \eqno{(5)}
$$

\section{Evaluation}

\par A key advantage of the Diffusion Co-policy is its generative quality; however, the model's effectiveness on human-robot collaborative tasks remains unexplored. The collaborative carrying task itself poses several interesting questions for evaluation: 1) Can it learn to effectively condition on obstacles? 2) Can it learn to effectively condition on its partner's behaviors and shared task representation? 3) Can it mutually adapt with real humans in test time? To attempt to address these questions, we compare our diffusion co-policy against several state-of-art imitation learning methods on a suite of evaluations described in the following sections. 

\par We select three learning-based methods for comparison:
\begin{itemize}
    \item \textbf{BC-LSTM-GMM \cite{mandlekar2021matters}}: Several works in \edits{HRI} have leveraged \edits{a} variant of this method, notably for the GMM output layer. We adapt the implementation from \cite{mandlekar2021matters} \edits{and did not condition inputs on past human actions; doing so led to worse performance on the task.}
    \item \textbf{VRNN Planner \cite{ng2022takes}}: This sampling-based planner autoregressively predicts team waypoints learned from demonstrations in receding horizon and does not condition on human actions.
    \item \textbf{Co-GAIL \cite{wang2022co}}: Co-GAIL learns collaborative behaviors from demonstrations and maps human behaviors to a latent space, which is then used to train a co-policy with Generative Adversarial Imitation Learning (GAIL).    
\end{itemize}

\par \edits{Some baselines do not incorporate map information. To improve task performance, we trained them using the augmented state representation, $s_t'$.} We also compare two variants of the Diffusion Co-Policy: one with past human action conditioning (\textbf{CoDP-H}), and the other without (\textbf{CoDP}).

\begin{figure}[t]
  \centering
  \vspace{3mm}
  
  \includegraphics[width=.8\linewidth]{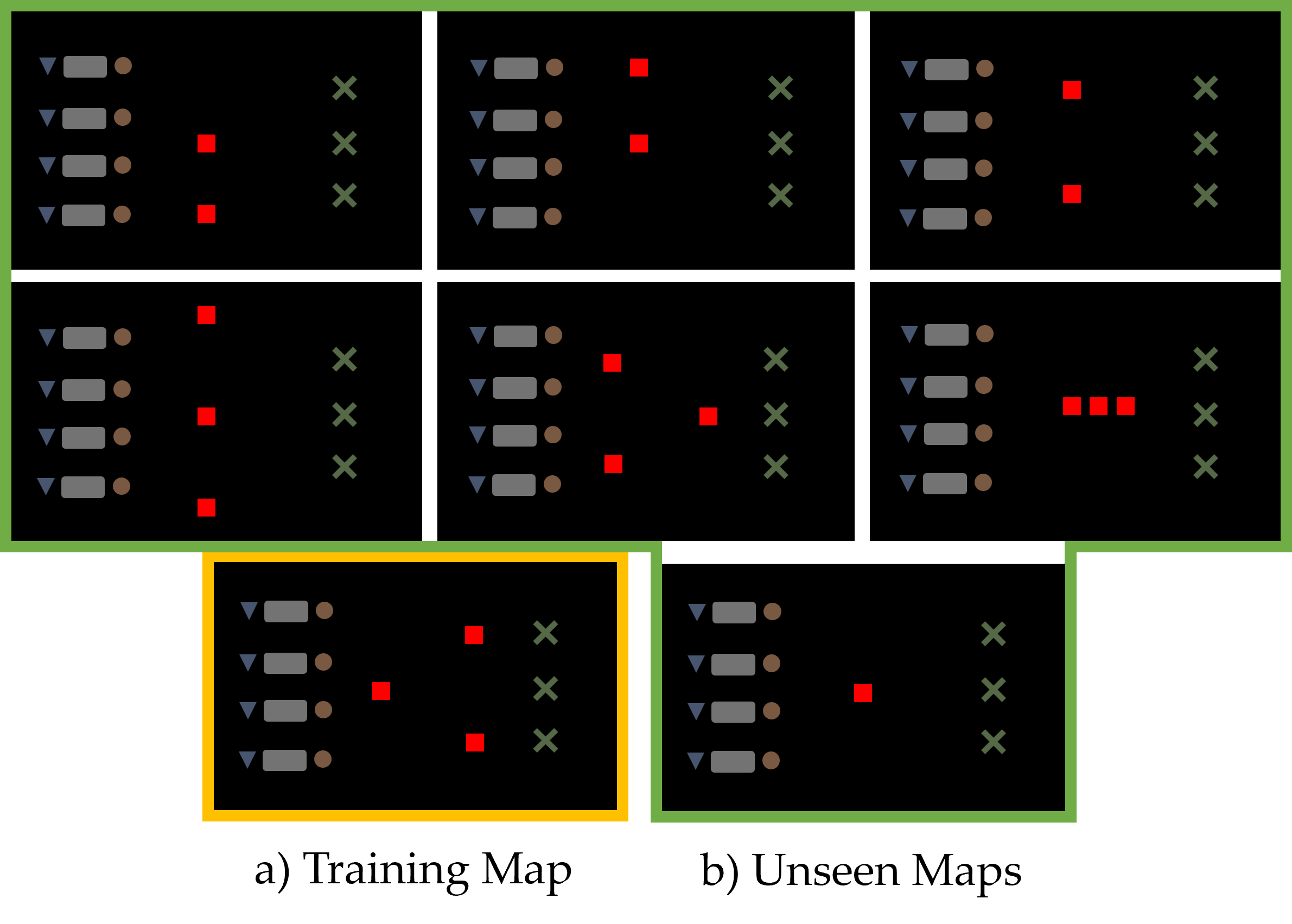}
  \caption{Co-planning evaluation training (yellow box) and test (green box) maps. Start poses and goal positions (green ``X"s) are shown, along with obstacle layouts (red boxes).}
  \label{fig:maps}
  \vspace{-3mm}
  
\end{figure}

\subsection{Experimental Setup}
\par  We trained the diffusion co-policy, CoDP, and variant (CoDP-H) to output joint actions at 10Hz, which were executed on a simulator running at 30Hz, applying a zero-order hold of 3 time steps for each planned. All other baselines were trained to produce outputs at 30Hz. \edits{We conduct the following experiments and user studies} for the collaborative carrying task to address the questions posed \edits{above}.
\subsubsection{\textbf{Co-planning \textit{(\edits{in simulation environment})}}} To test each method's ability to complete the task without a human in the loop and learn a representation of the map without the potential added benefit of human co-piloting corrections, we varied out-of-distribution obstacle locations while keeping the same distribution of initial and goal states from the training data. We executed $T_a = 8$ actions sampled with 100 denoising steps before replanning, which takes roughly 0.3 sec on a NVIDIA 3090Ti GPU. 
\subsubsection{\textbf{Human-in-the-loop evaluation \textit{(\edits{in simulation environment})}}}  In this \edits{user study}, the robot policies complete the task with a real human-in-the-loop, in various out-of-training-distribution settings. The human teleoperates the orange circle agent in the simulation using joystick control. The sampling scheme for the human-in-the-loop simulation evaluation is different than the co-planning setting due to having a human in the loop. To account for visual latency and reaction time, we execute $T_a = 1$ sampled actions with 34 inference steps to allow for planning time of roughly 0.1 sec, and zero-order hold each planned action for 3 time steps before executing the next planned action, resulting in low visual latency in the simulator. 
\par \edits{Here, we study the effect of the policy or planning method on success rate (i.e. the percentage of trials in which the user succeeds in completing the task with the robot). Our \textit{hypothesis} is that the diffusion policies will enable the robot to complete the task with a human at a higher success rate than the other methods, particularly since the diffusion policies are generating multimodal, multi-step predictions.}
\subsubsection{\textbf{Human-in-the-loop evaluation \textit{(\edits{in real environment})}}} In this \edits{user study, a human uses a joystick to teleoperate an Interbotix Locobot that is pin-connected via a rigid rod to another Locobot, operated by a policy or planner. We use the same policies trained in simulation and deploy them in the real environment in zero-shot sim-to-real transfer. We also use the same sampling scheme from the human-in-the-loop evaluation in simulation. To address the sim-to-real gap of the state space, we mapped a space in the experiment room and scaled it to the simulation environment coordinates, using data from motion capture. For this experiment, we consider two initial configurations: in the first, which we denote as ``Human Front", the human is placed closest to an obstacle such that they are inclined to make the decision to go above or below the obstacle before the robot; and in the second, which we denote as ``Robot Front", the robot is placed closest to an obstacle, implicitly forcing the robot to make the same decision before the human.}
\par \edits{In this study, we investigate the robustness of the robot's policy or planner to its initial configuration by determining whether there is a significant interaction effect of those two factors. Our \textit{hypothesis} is that robots running the diffusion methods should not see an interaction effect with initial configuration since diffusion methods can express a high degree of multimodality.}

\begin{figure}[t]
  \centering
  \vspace{2mm}
  \begin{subfigure}[t]{.48\columnwidth}
    \centering
    \includegraphics[width=\linewidth]{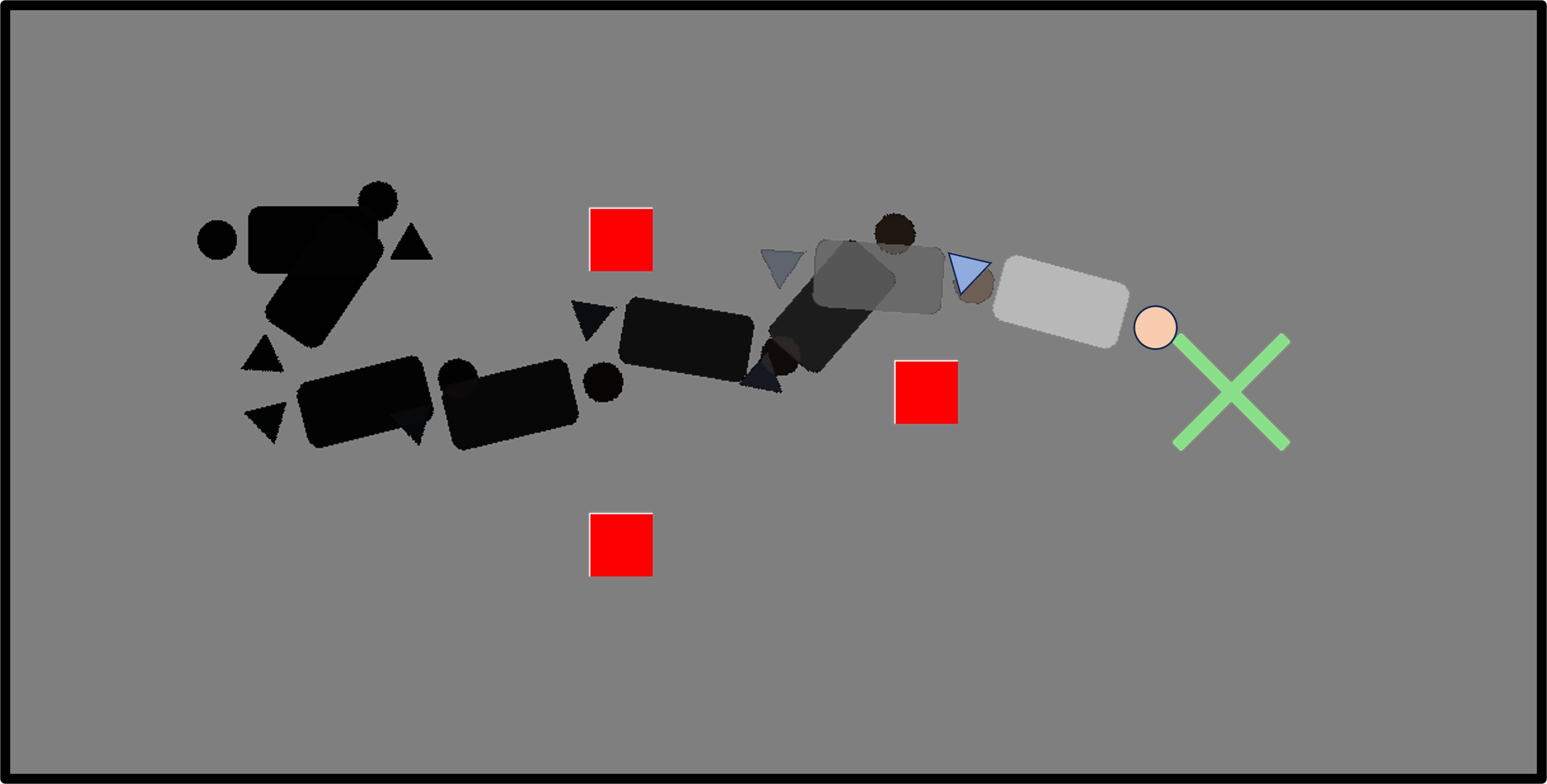}
    \caption{Mutual adaptation and shared task understanding with CoDP-H.}
    \label{fig:mutual_adapt}
  \end{subfigure}%
  \hfill
  \begin{subfigure}[t]{.48\columnwidth}
    \centering
    \includegraphics[width=\linewidth]{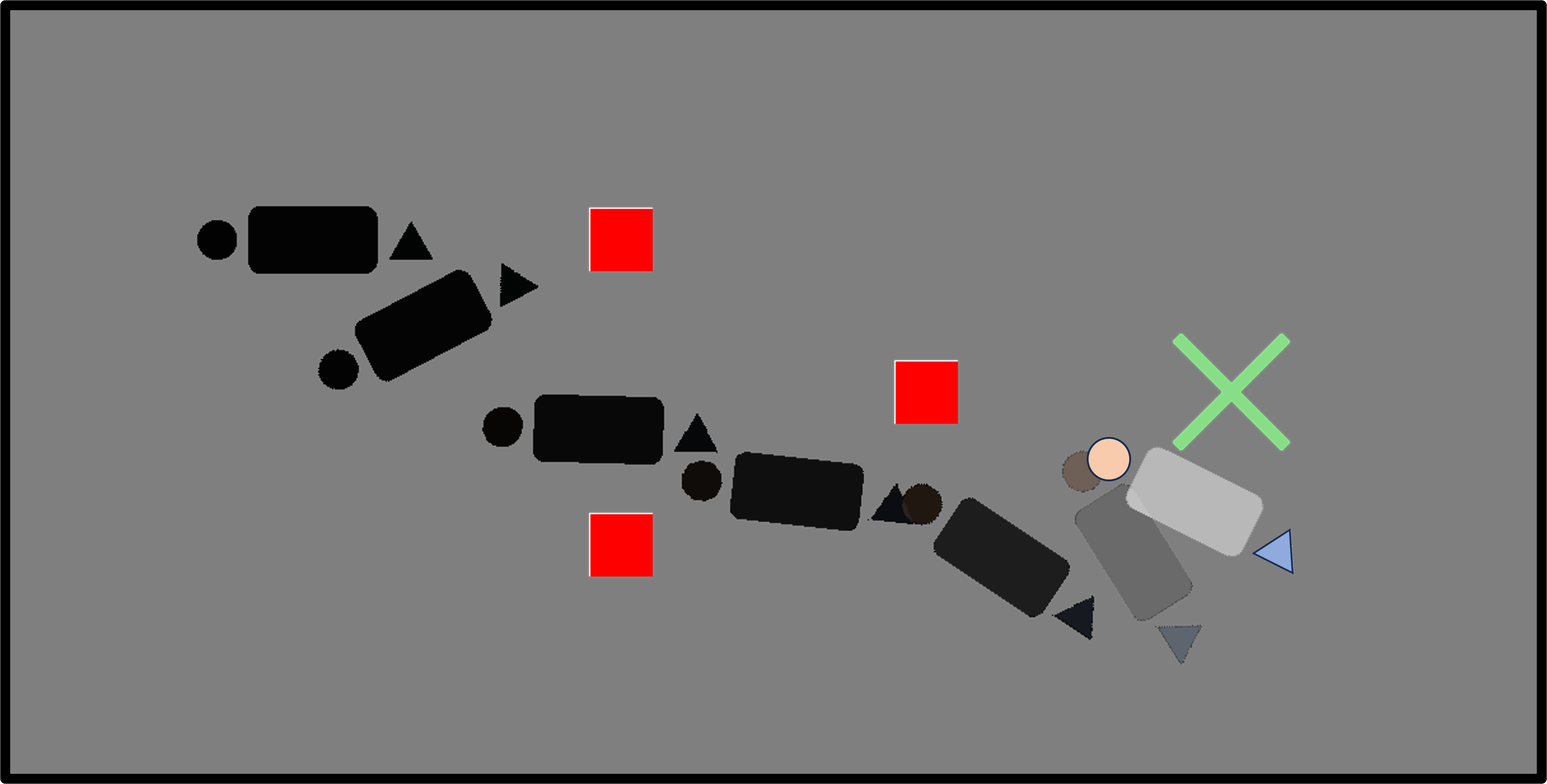}
    \caption{Leadership switching with CoDP-H.}
    \label{fig:leadership_switch}
  \end{subfigure}%
  \vskip\baselineskip
  \vspace{-1mm}
  \begin{subfigure}[t]{.48\columnwidth}
    \centering
    \includegraphics[width=\linewidth]{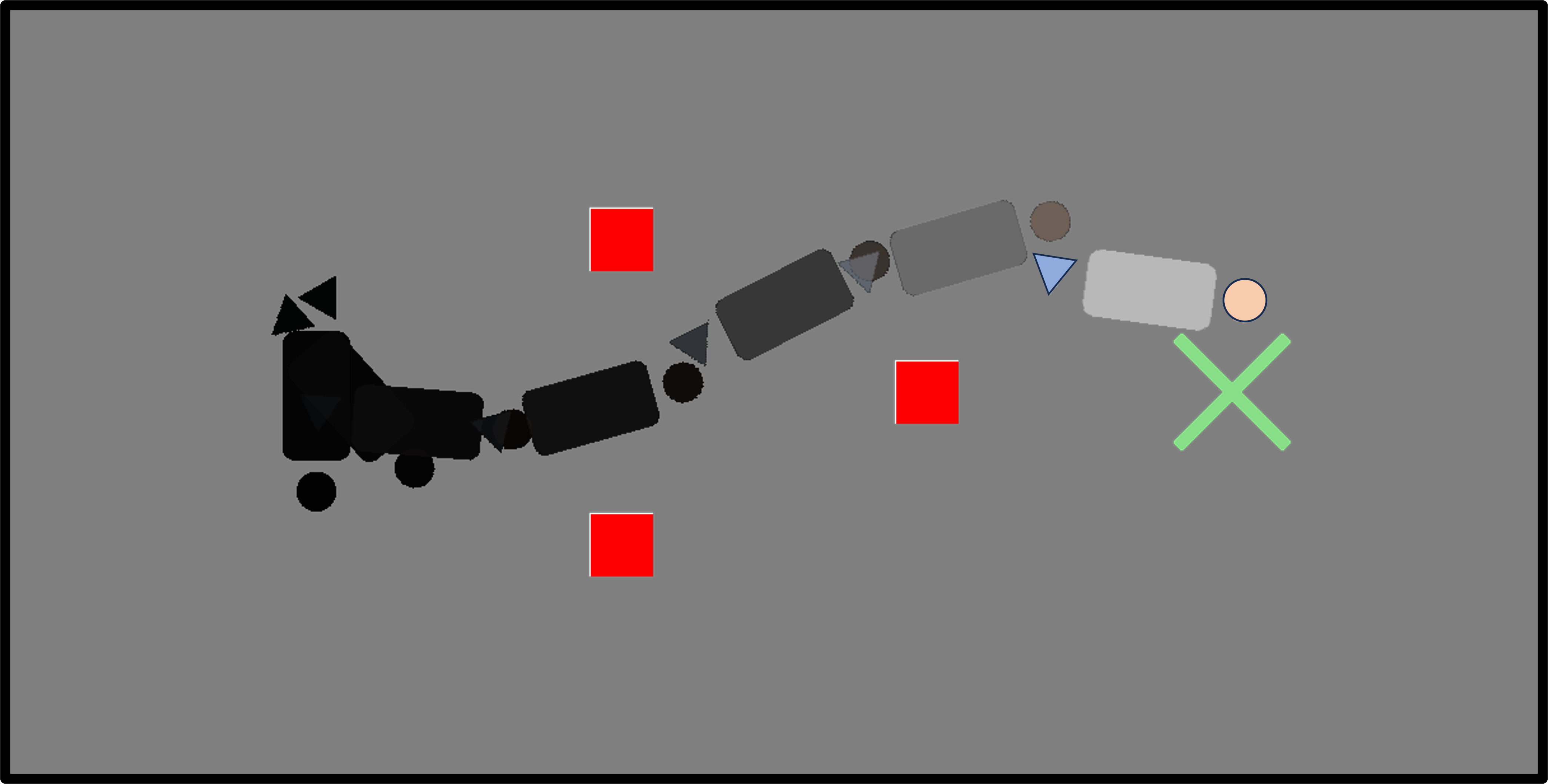}
    \caption{No human action conditioning (CoDP).}
    \label{fig:no_hcond}
  \end{subfigure}%
  \hfill
  \begin{subfigure}[t]{.48\columnwidth}
    \centering
    \includegraphics[width=\linewidth]{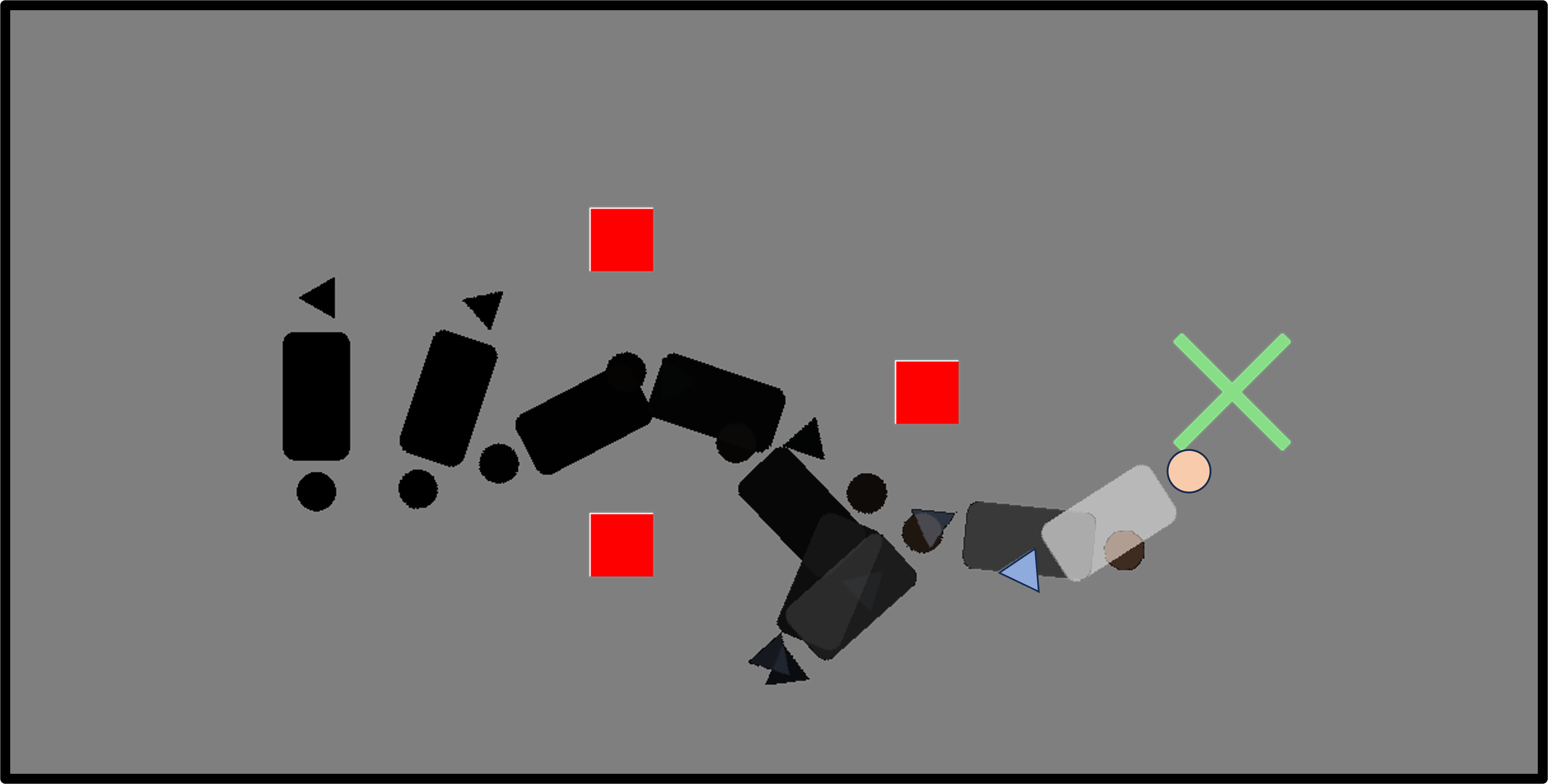}
    \caption{Learned partner behaviors with human action conditioning (CoDP-H).}
    \label{fig:hcond}
  \end{subfigure}%
  
  \hfill
  \caption{\textbf{Qualitative Observations:} (a) \textit{Mutual adaptation and shared task understanding} between human (orange circle) and CoDP-H agent (blue triangle).  (b) \textit{Leadership switching} with CoDP-H. (c) \textit{Without conditioning on past human behavior}, robot behavior is notably affected: the human becomes de-facto leader as the robot displays passive behavior. (d) \textit{With conditioning} on its partner's past actions, the robot actively takes the lead, and displays interesting leadership switching behaviors via pivoting.}
  \vspace{-4mm}
\end{figure}

\subsection{Dataset and Map Details}
\subsubsection{Training data} 
\begin{figure*}[!h]
  \centering
  \vspace{5mm}
  \adjincludegraphics[width=1.\textwidth, trim={{0.03\width} {0.04\height} {0.01\width} {0.15\height}}, clip]{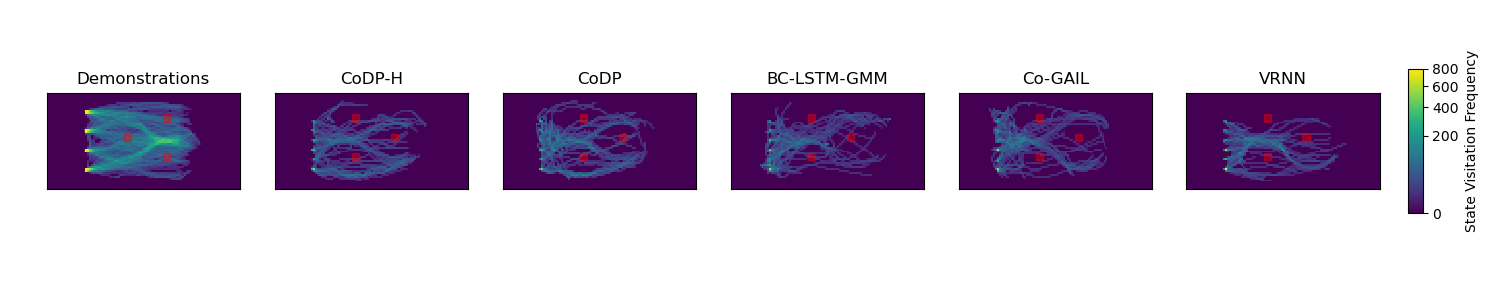}
  \vspace{-13mm}
  \caption{Heat map visualization of \textit{state visitation frequencies}, from left to right, of human-human demonstration data for training the models, and the human-in-the-loop simulation evaluations on novel, unseen maps for each policy or planner. In each scenario, one to three obstacles are placed in the the potential obstacle locations shown as red squares for each map.}
  \label{fig:state_visit}
  \vspace{-3mm}
\end{figure*}

The training dataset consists of 376 human-human demonstrations (179,993 environment interactions) on the collaborative carrying task \edits{collected by 5 distinct pairs of people} on a total of 36 possible configurations. Due to the added complexity of obstacle representation learning, cost of demonstration collection, and requirement of hand-engineered obstacle placement to allow for multimodal behaviors, we used up to a maximum of three obstacles in each training demonstration, with each initial, goal, and obstacle location illustrated in Fig. \ref{fig:maps}a. Note that while many offline dataset learning methods \cite{fu2020d4rl} augment data with trained RL policies, planners paired with PID controllers, etc., we recognize that such augmented data could skew our dataset distribution, particularly if these methods do not contain demonstrations of multimodal, sub-optimal, and inconsistent, yet ``human-like" behaviors. For example, RRT planners do not exhibit the same behaviors (e.g. rotations, distance from obstacles) as human demonstrators on the collaborative carrying task \cite{ng2022takes}. \edits{The demonstration data in this work contains multimodal behaviors, as seen in Fig. \ref{fig:state_visit}.}

\subsubsection{Test maps} For the co-planning evaluation, we evaluated on all possible combinations of unseen map settings outlined in Fig. \ref{fig:maps}b. For the human-in-the-loop simulation evaluation, we sampled from a subset of unseen maps, goals, in addition to different initial orientations. For reference, $\pi$ is the initial table orientation depicted in Fig. \ref{fig:maps}, and we included four total initial orientations in our sampling, i.e. $[0, \frac{\pi}{2}, \pi, \frac{3\pi}{2}]$. We then evaluated on the same sampled subset per method. On the real robot evaluation, we used a one obstacle unseen map with two initial orientations: robot facing the obstacle first, and human facing the obstacle first. 


\begin{table}[h]
\centering
\vspace{2mm}
\resizebox{1.0\columnwidth}{!}{%
\begin{tabular}{|c|c |c |}
\hline
Method & \thead{Unseen Maps \\ Success (\%)} & \thead{Test Holdout Maps \\ Success (\%)}  \\
\hline
 CoDP-H & \textbf{40.48} / \textbf{32.54} & \textbf{78.57} / \textbf{77.38}  \\

 CoDP & 32.14 / 28.97 & 72.62 / 67.86  \\

 BC-LSTM-GMM\cite{mandlekar2021matters} & 19.05 / 17.06 & 60.71 / 55.16 \\

 Co-GAIL\cite{wang2022co} & 25.00 / 22.22 & 47.62 / 39.28 \\

 VRNN\cite{ng2022takes} & 22.62 / 19.44 & 22.62 / 20.24 \\
\hline

\end{tabular}
}
\caption{\textbf{Co-planning Results.} Reported success rates for each method as (max performing seed / average performance over 3 random training seeds) over a total of 84 randomly selected test holdout maps (Fig. \ref{fig:maps}a) and 84 novel configurations on unseen maps (Fig. \ref{fig:maps}b). Our results show that the diffusion co-policy conditioned on past human partner actions (CoDP-H) outperformed all state-of-art imitation learning methods and baselines on the co-planning evaluation.}
\label{table:coplanning_results}

\vspace{-5mm}
\end{table}

\begin{figure*}[!h]
  \centering
  \vspace{1mm}
  \adjincludegraphics[width=1.\textwidth, trim={{0.02\width} {0.02\height} {0.00\width} {0.05\height}}, clip]{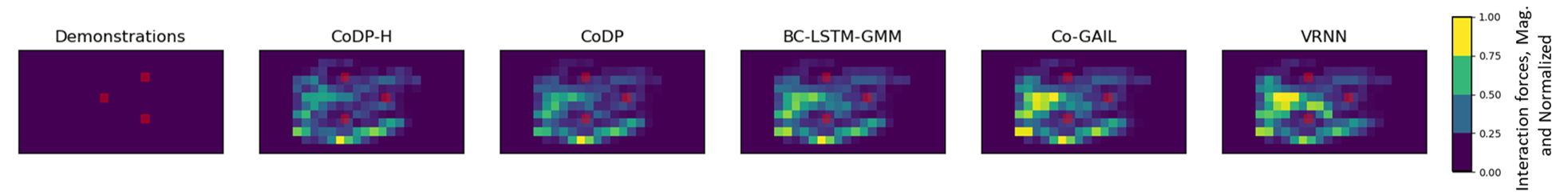}
  \vspace{-6mm}
  \caption{\edits{Comparison of interaction forces from the human-human demonstrations and simulation evaluations with a human-in-the-loop on novel, unseen maps. For each trajectory, we binned the interaction force \textit{magnitudes} (see color-map bar for bins) at each location in the discretized map, then \textit{normalized} the magnitudes across \textit{all} trajectories. Each map shows $1-3$ obstacles at potential obstacle locations (red squares). \edits{The diffusion methods show overall} lower magnitudes of interaction forces across all human-in-the-loop evaluations in simulation.}}
  \label{fig:interf_binned}
\vspace{-4mm}
\end{figure*}

\subsection{Simulation Results: Key Findings} 
\subsubsection{Co-planning} 
We subjected the co-policies to the collaborative carrying task on maps seen in training, as well as maps unseen in training. \textbf{CoDP-H} consistently outperforms other baselines on all maps tested (Table \ref{table:coplanning_results}). Considering the few maps and obstacle configurations used during training, methods leveraging diffusion models exhibited unexpectedly high performance on maps with novel obstacle locations, suggesting that the diffusion co-policy was able to learn obstacle representations efficiently.  

\subsubsection{Human-in-the-loop trials}
\par \edits{The diffusion methods} outperform other baselines on the human-in-the-loop evaluation on task success rate (Table \ref{table:hil_sim_results}). \edits{We validate this trend by conducting a one-way repeated measures ANOVA to examine the effect that the methods had on task success rate. Results showed statistically significant differences in success rate (F(4, 16) = 11.65, p~$<$~.001). We ran a post-hoc analysis with Tukey HSD corrections for multiple comparisons, which showed the human-in-the-loop performance of the diffusion methods to be significantly different from that of Co-GAIL, BC-LSTM-GMM, and VRNN, and all contrasts with p~$<$~.0001. The contrasts support our hypothesis that using a diffusion network for the robot policy yields higher success rates on the collaborative human-robot task, as it surpasses other methods at predicting multimodal behavior.} Fig. \ref{fig:state_visit} shows
heat maps of the state visitation frequencies for the human-in-the-loop simulation evaluations for each policy or planner, as well as rollouts from the human-human demonstration dataset. Note in the test maps that obstacles were placed in locations where the table frequently visited in the training data. This suggests that the diffusion methods are able to learn a shared task representation and perform well on unseen maps with a real human partner.
\par Table \ref{table:hil_sim_results} also demonstrates the planning time disadvantage of diffusion-based methods; yet, despite planning less frequently and requiring interpolation methods, diffusion-based methods achieve higher success rate on the task. 

\begin{table}[h]

\centering
\resizebox{1.0\columnwidth}{!}{%
\begin{tabular}{| c | c | c | c |}
\hline
Method & \thead{Success \\ (\%)} & \thead{Time \\ (s)} & \thead{Plan Time \\ (ms)}  \\
\hline
 CoDP-H & \textbf{68 $\pm$ \edits{2.1}} & 32.3 $\pm$ \edits{0.2} & 93.5 \edits{$\pm$ 4.0} \\

 CoDP  & 61 $\pm$ \edits{3.7} & 31.0 $\pm$ \edits{0.1} & 87.0 \edits{$\pm$ 3.8}  \\

 BC-LSTM-GMM\cite{mandlekar2021matters}  & 36 $\pm$ \edits{1.4} & 22.3 $\pm$ \edits{0.5} & 0.7\edits{45} \edits{$\pm$ 0.004}  \\

 Co-GAIL\cite{wang2022co}  & 37 $\pm$ \edits{1.9} & 23.9 $\pm$ \edits{0.9} & \textbf{0.26\edits{7} \edits{$\pm$ 0.003}}   \\

 VRNN\cite{ng2022takes}  & 35 $\pm$ \edits{3.6} & \textbf{15.8 $\pm$ \edits{1.4}} &  18.52\edits{$\pm$ 0.02}  \\
\hline

\end{tabular}
}
\caption{\textbf{Human-in-the-Loop Simulation Results.} \edits{The max performing model seed was used for each robot planner or policy, which played with human subjects ($N=5$) for a total of 60 randomly selected configurations on unseen maps (Fig. \ref{fig:maps}b). Standard error (SE) is reported for all measurements, including success rate $(\%)$, time to completion (s) for successful trajectories, and average time for the model to plan (ms).} Our results show that the diffusion co-policy conditioned on past human partner actions (CoDP-H) outperformed all baselines for the human-in-the-loop evaluation in simulation.}
\vspace{-5mm}
\label{table:hil_sim_results}
\end{table}

\subsection{Interesting Behaviors in Human-in-the-loop Evaluation} 
Diffusion co-policy demonstrated interesting collaborative behaviors on novel configurations in simulation evaluation. We highlight them qualitatively as follows:

\subsubsection{Mutual adaptation and shared task understanding} 
Fig. \ref{fig:mutual_adapt} demonstrates an instance of mutual adaptation as well as shared task understanding. Initially, both agents simultaneously \edits{choose} different strategies: the robot (blue triangle) applies a downward force while the human (orange circle) applies an upward force, resulting in an in-place moment on the table. The human leads, while the robot maintains awareness of obstacles and rotates to avoid them, demonstrating shared task understanding between human and robot. These behaviors lend insight into the better performance of CoDP-H in human-in-the-loop evaluations. 

\subsubsection{Leadership switching} 
Fig. \ref{fig:leadership_switch} demonstrates an instance of leadership switching, which occurs several times over the course of the trajectory. The human starts leading by moving below the obstacle, but the robot takes over by maintaining its lead in front. Both agents approach the goal past the final obstacle. This demonstrates the ability to switch roles while maintaining task understanding.   

\subsubsection{Learning partner behaviors} 
Conditioning on past partner actions allows the robot to develop a better understanding of the task and its partner's intentions. Without this past action conditioning, the robot acts passively, leaving the human to lead (Fig. \ref{fig:no_hcond}). Fig. \ref{fig:hcond} shows that the CoDP-H robot is capable of \edits{pivoting, a} proactive behavior, since it has learned to associate past partner actions with observations. \edits{This behavior was not seen in the demonstration data.} 

\subsubsection{Low interaction forces} 
\begin{figure*}[!h]
  \centering
  \adjincludegraphics[width=.8\textwidth, trim={{0.00\width} {0.0\height} {0.01\width} {0.00\height}}, clip]{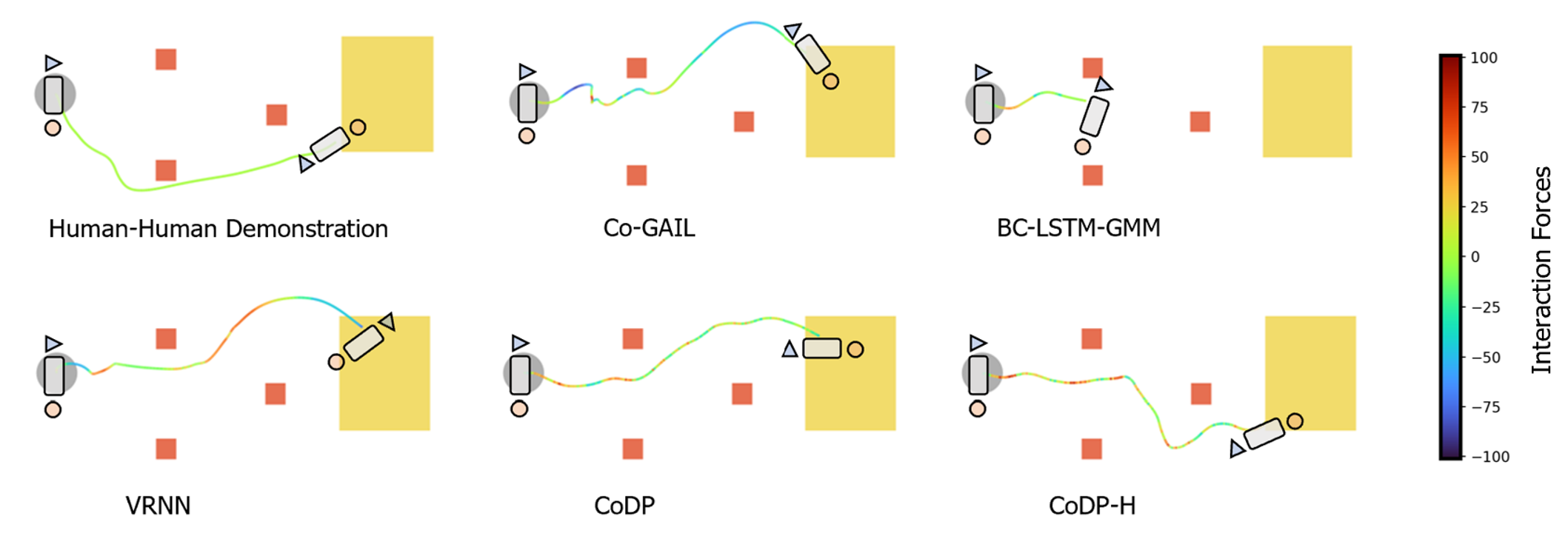}
  \caption{Visualization of interaction forces over rollouts on a sample map configuration played between each robot policy or planner (blue triangle) with a human (orange circle) for evaluation in simulation, with the goal (yellow region) and obstacles (red squares) shown. Closer to zero is better, as demonstrated in the human-human exemplar. The diffusion policies show overall lower (in magnitude) interaction forces over the course of the trajectory.}
  \vspace{-3mm}
  \label{fig:interf}
\end{figure*}

Stretching or compressing may occur during transport of an object, indicating non-zero interaction forces \cite{kumar1988force}. Interaction forces do not contribute to motion, and can lend insight into periods of collaboration, disagreement, and other decision points in the trajectory. \edits{Fig. \ref{fig:interf_binned} shows a 2D heat map of normalized interaction force magnitudes over all trajectories for each method.} Interaction forces less than 0.25 are generally negligible, and all human-human demonstrations displayed a negligible frequency of non-zero interaction forces. \edits{Forces} between 0.25 - 0.75 may indicate a decision point or dissent; \edits{those} above 0.75 \edits{are} strong indicators of dissent or human corrective action. Fig. \ref{fig:interf} shows interaction forces across an example map configuration for each method with a human in the loop. Across all methods, \edits{the diffusion policies display lower interaction forces over most areas in the maps}.

\begin{table}[h]
\centering
\resizebox{0.9\columnwidth}{!}{%
\begin{tabular}{|c|c |c |}
\hline
Method & \multicolumn{1}{c|}{\thead{Human Front \\ Success (\%)}} & \multicolumn{1}{c|}{\thead{Robot Front \\ Success (\%)}} \\
\hline
 CoDP-H & \textbf{75.0 $\pm$ 4.6} &  91.7 $\pm$ 3.4 \\

 CoDP & 66.7 $\pm$ 4.3 &  91.7 $\pm$ 3.4 \\
 BC-LSTM-GMM\cite{mandlekar2021matters} & \edits{33.3} $\pm$ 4.3 & \textbf{100.0 $\pm$ 0.0} \\

 Co-GAIL\cite{wang2022co} & 66.7 $\pm$ 4.3 & 50.0 $\pm$ 7.4 \\
 VRNN\cite{ng2022takes} & \textbf{75.0 $\pm$ 4.6} & 75.0 $\pm$ 7.0 \\
\hline

\end{tabular}
}
\caption{\textbf{Human-in-the-loop Real Robot Experimental Results.} We tested on a real-robot scenario with a single centered obstacle (see bottom unseen map in Fig. \ref{fig:maps}), in two initial configurations: one with the human facing the obstacle first (``human front"), and the other with the robot facing the obstacle first (``robot front"). Average success rate (\%) is reported with std. dev. over subjects ($N=6$), with a total of 12 trajectories per initial configuration. BC-LSTM-GMM outperforms CoDP-H in the robot front setting, but does significantly worse in the human front setting.}
\vspace{-5mm}
\label{table:real_robot_results}
\end{table}

\subsection{Real robot evaluation}

\par
\begin{figure*}[!h]
  \centering
  \vspace{1mm}
  \adjincludegraphics[width=1.\textwidth, trim={{0.0\width} {0.0\height} {0.0\width} {0.0\height}}, clip]{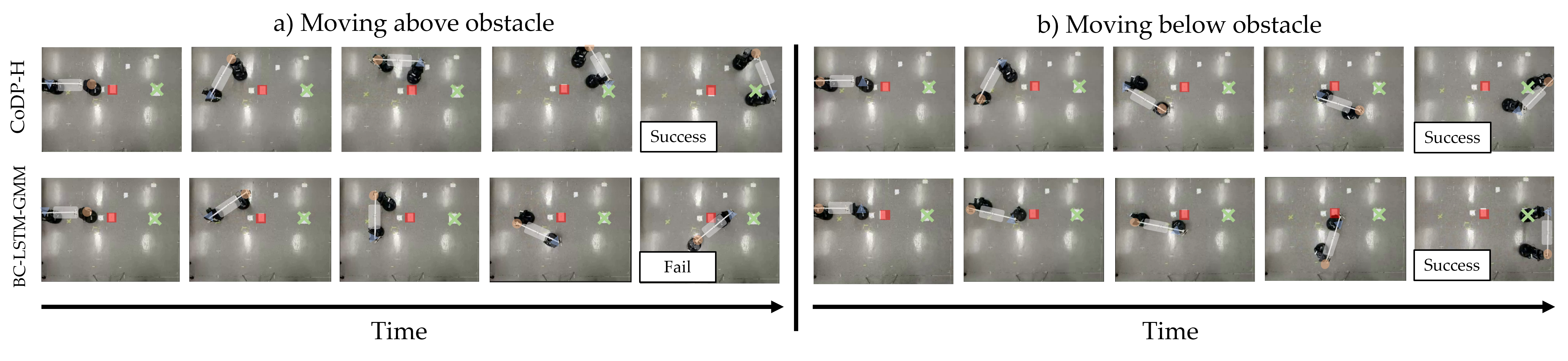}
  \vspace{-6mm}
  \caption{All robots are represented by the blue triangle. The human is the orange circle, the goal is a green "X", and the obstacle is a red square. BC-LSTM-GMM prefers a route \textit{below} the obstacle, leading to greater success rate if the human also chooses the route below the obstacle; otherwise, it can lead to failure. With the BC-LSTM-GMM robot facing the obstacle first, the human serves a more passive role, resulting in a higher success rate compared to when the human faces the obstacle first, since the robot will choose the route below the obstacle.}
  \label{fig:real_demo}
\vspace{-3mm}
\end{figure*}

 \edits{Table \ref{table:real_robot_results} shows that the diffusion methods outperform the other baselines, except for BC-LSTM-GMM in the ``Robot Front" case.} While BC-LSTM-GMM appears to outperform the diffusion methods in the ``Robot Front" case, it performs poorly in the ``Human Front" case. BC-LSTM-GMM prefers a route below the obstacle; if the human partner happens to adapt to the robot or pick the same route, this tends to lead to success. However, unlike CoDP-H, it is unable to adapt to move above the obstacle when necessary to achieve task success, as seen in Fig. \ref{fig:real_demo}. \edits{This suggests a significant interaction between the policy or planner method and the initial configuration of the robot with respect to the obstacle, which is confirmed by results from a two-way repeated measures ANOVA for interaction effects, \mbox{F(4,20)~=~3.170}, p~=~0.036. Main effects on task success rate were also significant for initial configuration, F(1, 5)~=~7.857, p~=~0.038, and for method, F(4, 20)~=~3.152, p~=~0.037. We further investigated the interaction effect of the initial configuration for each method. Adjusted P-values using Holm multiple testing corrections show that the effect of initial configuration on success rate was significant for BC-LSTM-GMM (p~=~0.001), but not for the other methods. This supports our hypothesis, and suggests that BC-LSTM-GMM is affected by the initial configuration and therefore less robust to multimodal outcomes that arise from human-robot interactions. While the other methods do not show significant interaction effects, they perform poorly in the task compared to the diffusion methods.}

\vspace{-1mm}

\section{Conclusion}

\par In this work, we explore using action predictions from diffusion models to plan collaborative actions that synergize well with real humans in the loop. We show that a co-policy developed with a Transformer-based diffusion model and conditioned on past human actions can not only plan multimodal action sequences with real humans-in-the-loop to achieve high success rates, but also qualitatively display compelling collaborative behaviors in novel, out-of-training-distribution settings, including mutual adaptation, shared task understanding, and leadership switching.
\par Our study has several limitations. \edits{The time required to generate a sample with the diffusion policy} is longer than other \edits{methods}. We also faced several limitations in the real robot experiments, including physical capabilities of the robots, physical space constraints, and human subject variance. \edits{To extend this method for a co-manipulation task similar to \cite{sirintuna2022human}, a dataset with tactile feedback from both agents co-manipulating a table would be highly beneficial for the learned task representation.} Despite limitations in this work, the diffusion co-policy has demonstrated the significance of an expressive policy for human-robot collaboration, i.e. one that can capture a high degree of multimodality, predict actions in a temporally consistent manner, and recognize a wide range of frequencies of actions in order to seamlessly integrate with a human.


\bibliographystyle{unsrt}
\bibliography{refs}

\end{document}